\documentclass{article}

\usepackage[eandd, final]{neurips_2026}

\usepackage[utf8]{inputenc} 
\usepackage[T1]{fontenc}    
\usepackage{hyperref}       
\usepackage{url}            
\usepackage{booktabs}       
\usepackage{amsfonts}       
\usepackage{nicefrac}       
\usepackage{microtype}      
\usepackage{xcolor}         
\usepackage{multirow}
\usepackage{graphicx}
\usepackage{amsmath}

\newcommand{\sym}[1]{\rlap{$^{#1}$}}
\title{Scaling Point-in-Time Language Models:\\ Enabling Economic Evaluation of Embeddings }

\author{%
  Bryan Kelly \\
  Yale School of Management\\
  Yale \\
  New Haven, CT 06511 \\
  \texttt{bryan.kelly@yale.edu} \\
    \And Semyon Malamud\\
      Swiss Finance Institute\\
  EPFL\\
  Lausanne, Switzerland. \\
  \texttt{semyon.malamud@epfl.ch} \AND Johannes Schwab\\ 
Swiss Finance Institute\\
  EPFL\\
  Lausanne, Switzerland. \\
  \texttt{johannes.schwab@epfl.ch} \\
  \And Teng Andrea Xu \\
    AQR Capital Management \\
  Greenwich, CT 06830 \\
  \texttt{andrea.xu@aqr.com} \\
}

\begin{document}

\maketitle

\begin{abstract}
Large language models trained on unrestricted internet corpora inevitably embed information from the future, introducing lookahead bias that compromises the validity of backtests and causal inference in finance and the social sciences. Point-in-time language models—trained exclusively on text available up to each calendar date—eliminate this leakage by construction, but existing efforts typically produce models that lag substantially behind their unconstrained counterparts. We show that this performance gap can be narrowed through scale. Training decoder-only transformers with up to 4 billion parameters on 1 trillion chronologically filtered tokens from FineWeb, we construct a sequence of yearly model checkpoints spanning 2013–2024. Across a range of common-sense reasoning and language understanding benchmarks, our models approach the performance of leading open-weight models of comparable size (such as Gemma-3-4B and LLaMA-7B) trained on temporally unrestricted data, although a performance gap remains on several tasks. Finally, in a strict out-of-sample economic evaluation task, portfolios built from point-in-time embeddings achieve robust positive Sharpe ratios and perform close to full-sample counterparts that violate temporal validity, indicating that chronologically consistent language models can extract economically meaningful signals without relying on look-ahead bias. We release the complete pipeline—including dataset construction, training infrastructure, and evaluation code—to enable reproducible point-in-time language modeling and to support research applications that require strict temporal validity. Code is available publicly on \href{https://github.com/DjoFE2021/Point-In-Time-LLM}{GitHub} and model checkpoints as well as training datasets can be accessed on \href{https://huggingface.co/Diamegs}{Hugging Face}.
\end{abstract}

\section{Introduction}
\label{sec:intro}


Large language models are increasingly used in economics, finance, and the social sciences to extract structured signals from unstructured text \citep{gentzkow2019text, korinek2023language, horton2023large}. These models enable new approaches to measuring expectations, analyzing narratives, and forecasting economic outcomes.
However, current research largely relies on foundation models trained on temporally unrestricted data \citep{brown2020language, touvron2023llama, team2025gemma, bi2024deepseek}. Because these models are trained on the entire internet corpus simultaneously, they inevitably encode knowledge from future time periods relative to the historical data being analyzed.
This phenomenon, commonly referred to as \emph{lookahead bias} or temporal leakage~\citep{glasserman2023, lopez2023can,  sarkar2024lookahead, levy2024, ludwig2025, huang2026aierrors}, can invalidate empirical results. In financial applications, such bias can distort estimates of risk premia and invalidate backtests. In economic forecasting, it can lead to artificially optimistic predictions by implicitly incorporating future information.

A natural remedy is to train \emph{point-in-time} language models: models whose training data are restricted to text published on or before a given date. Recent work has demonstrated the feasibility of this approach, first with~\citep{he2025chronologically, he2025instruction}  (ChronoBERT and ChronoGPT), then followed by~\cite{yan2026datedgpt} (DatedGPT), producing chronologically consistent models that outperform earlier no-leakage baselines. However, these models remain substantially smaller and weaker than leading open-weight models, raising a practical question: \emph{how much performance must one sacrifice to eliminate lookahead bias in large language models?}

We show that the answer is: very little. By scaling both model size and training data---up to 4 billion parameters and 1 trillion tokens of chronologically filtered web text---we obtain point-in-time models whose common-sense reasoning and language comprehension approach those of Gemma-3-4B~\citep{team2025gemma} and LLaMA-7B~\citep{touvron2023llama}, models trained on full, temporally unrestricted corpora. Moreover, the economic forecasts produced by our point-in-time language models are free from lookahead and produce tradable portfolios with strong out-of-sample mean returns. 

Our results suggest that the performance gap attributed to temporal restrictions is largely a gap in scale~\citep{henighan2020scaling,kaplan2020scaling, hoffmann2022training}, not an inherent limitation of the point-in-time paradigm.

\textbf{Contributions.} Our contributions are fourfold. \emph{First}, we advance the state of the art in point-in-time LLM training. Starting from the GPT-2-based architecture of~\cite{modded_nanogpt_2024}, we scale training data by approximately $140\times$, increase model size from 1.5B to 4B parameters, extend the context length from 1,792 to 2,048 tokens, and increase the embedding dimension from 1,536 to 4,096. The resulting models achieve zero-shot accuracy on standard benchmarks that is within a few percentage points of leading open-weight models trained without temporal constraints. \emph{Second}, we instruction-tune our models using LoRA~\citep{hu2022lora}   and evaluate them on IFEval~\citep{zhou2023instruction}, a programmatically verifiable benchmark that avoids the well-documented biases of LLM-as-a-judge evaluation~\citep{zheng2024cheating, liu2023llms, shen2023large, wang2024large}. \emph{Third}, we release the complete project resources---covering dataset construction, model training, evaluation code, and yearly model checkpoints---thereby substantially lowering the barrier to reproducible point-in-time LLM research. \emph{Fourth}, and foremost, we assess the economic value of our models in an asset pricing application. We construct point-in-time textual signals from news and use them to forecast returns and risk premia. This setting provides a stringent out-of-sample test of whether temporally consistent language models extract economically meaningful information rather than inadvertently exploiting look-ahead bias. We show that point-in-time models deliver robust predictive performance and economically significant gains relative to models trained on standard, non-temporally constrained corpora, highlighting the importance of chronological consistency for financial applications.
\section{Literature Review}
\label{sec:literature}
\textbf{Text-based asset pricing.} A large body of literature studies how news text forecasts stock returns. Early work relies on dictionary-based or word-count methods~\citep{tetlock2007giving, tetlock2008more}, while supervised approaches extract sentiment signals tailored to return predictions~\citep{ke2019predicting}. More recently, large language models have enabled richer representations of news content. \cite{chen2022expected} show that LLM embeddings of news articles strongly predict next-day returns, and \cite{lopez2023can} demonstrate that prompting ChatGPT for headline sentiment yields robust trading signals. \cite{bybee2024business} use topic modeling on news text to construct interpretable macroeconomic factors. \cite{didisheim2026inefficient} build on these advances by decomposing news embeddings into predictable and surprise components, uncovering a ``pure news'' anomaly that exceeds many previously documented market anomalies. Crucially, they verify their findings using chronologically consistent LLMs from~\cite{he2025chronologically}, confirming that the anomaly is not an artifact of lookahead bias---precisely the kind of application that motivates our work.

\textbf{In-silico policy analysis.}  
A long-standing objective in economics and public policy is the ability to conduct counterfactual experiments on complex social systems. In practice, real-world policy experiments are costly, slow, and often ethically constrained. As a result, policymakers frequently rely on observational data and structural modeling assumptions. Recent work suggests that large language models may serve as synthetic agents for conducting \emph{in-silico experiments} that simulate the responses of individuals or institutions to hypothetical scenarios \citep{horton2023large}. However, the validity of such simulations depends critically on ensuring that models do not possess knowledge that would only become available in the future relative to the simulated time period. Point-in-time language models uniquely enable temporally consistent simulations. Policymakers could use these models to explore how populations might respond to policy changes, regulatory reforms, or economic shocks under realistic historical information environments.

\textbf{Point-in-time language models.} Our work builds on and extends the point-in-time LLM framework introduced by~\cite{he2025chronologically} and~\cite{he2025instruction}. In those papers, the authors train ChronoBERT and ChronoGPT, two families of chronologically consistent models with knowledge cutoffs beginning in 1999, and demonstrate competitive performance on NLP benchmarks and financial forecasting tasks. This line of work was recently extended in~\cite{yan2026datedgpt}. We demonstrate that this paradigm can be pushed substantially further through increased training data, model scaling, and  cutting-edge training techniques. Specifically, we scale up the original GPT-2-based ChronoGPT from a 1.5B-parameter model to a 4B-parameter decoder-only transformer trained on 1 trillion tokens, while extending the context length from 1,792 to 2,048 tokens and increasing the embedding dimension from 1,536 to 4,096 to align with architectures such as Mistral-7B~\citep{jiang2023mistral7b}. Because ChronoBERT is not a generative model, we use ChronoGPT and DatedGPT as the main benchmarks for the remainder of the paper.

\section{Methodology}
\label{sec:methodology}

The current literature suggests that large language models should be trained in stages. Models are first pre-trained on large-scale, general-purpose corpora to acquire broad linguistic and world knowledge~\citep{brown2020language, touvron2023llama}. They are then fine-tuned on instruction-following datasets to improve their generalization and zero-shot performance on unseen tasks~\citep{wei2021finetuned, sanh2022multitaskpromptedtrainingenables, ouyang2022traininglanguagemodelsfollow}. In this work, we closely follow the curriculum training procedure described in~\cite{lambert2024tulu}, with a new initial stage devoted to producing a strong pre-trained point-in-time base model. Our primary motivation is twofold: first, to demonstrate that point-in-time models can achieve common-sense reasoning and language comprehension on par with models trained on full, temporally unrestricted corpora; and second, to provide a fully reproducible training pipeline, including code and data configurations, to facilitate future research.

\textbf{Model.} We train decoder-only large language models with 1.5B and 4B parameters (henceforth PIT-1.5B and PIT-4B, respectively), adapting the implementation
of~\cite{modded_nanogpt_2024} to suit our requirements. The
architecture builds upon the GPT framework~\citep{radford2018improving} and incorporates several recent advances in optimization and scaling, including matrix function-based preconditioning~\citep{higham2008functions, schulz1933iterative}, learning rate modernization~\citep{bernstein2024old}, distributed
Shampoo optimization~\citep{gupta2018shampoo, anil2020scalable}, scaling strategies~\citep{haegele2024scaling}, and architectural
refinements such as value residual learning~\citep{zhou2024value}, as demonstrated in
Gemma~2~\citep{gemma2024gemma2}. The model is trained using the standard next-token prediction
objective~\citep{radford2018improving}. 
Under this autoregressive formulation, the joint probability of a
sequence $(x_1,\ldots,x_T)$ factorizes as
\[
p_\theta(x_1,\ldots,x_T)
=
\prod_{t=1}^{T} p_\theta(x_t \mid x_1,\ldots,x_{t-1}).
\]
In practice, the model minimizes the
negative log-likelihood (cross-entropy) of the next token:
\[
\mathcal{L}(\theta)
=
-\mathbb{E}_{x \sim \mathcal{D}}
\left[
\sum_{t=1}^{T}
\log p_\theta(x_t \mid x_1,\ldots,x_{t-1})
\right].
\]
The model is trained on a temporally ordered stream of tokens, directly aligning our setup with the incremental and continual learning literature~\citep{gupta2023continual, ke2023continual, parmar2024reuse, chen2025towards}. A complete description of the architecture and hyperparameters is provided in Appendix~\ref{app:training-config}.

\textbf{Pre-Training.} The selection of a pre-training corpus is a critical design decision, as an inappropriate data composition may lead to catastrophic forgetting~\citep{mccloskey1989catastrophic, kirkpatrick2017overcoming,luo2025empirical, kotha2023understanding}. We use the FineWeb dataset~\citep{penedo2024fineweb}, which provides a high-quality, filtered and deduplicated snapshot of internet text spanning from 2013 to 2025. The dataset comprises 15 trillion English tokens drawn from 96 Common Crawl~\citep{commoncrawl} snapshots, processed through language identification and quality filtering pipelines. Models trained on FineWeb have been shown to consistently outperform those trained on other publicly available web-scale corpora~\citep{DBLP:journals/jmlr/RaffelSRLNMZLL20, penedo2023refinedweb, soldaini-etal-2024-dolma, soboleva2023slimpajama, ortizsuarez:hal-02148693, gao2020pile} across a range of language understanding, reasoning, and knowledge benchmarks.

\textbf{Instruction Fine-Tuning.} Pre-training with a self-supervised next-token prediction objective on massive unlabeled corpora induces broad linguistic and world knowledge, but does not directly optimize performance on downstream tasks~\citep{raffel2020exploring}. Fine-tuning addresses this limitation and serves as a central mechanism of modern transfer learning in NLP, enabling pre-trained models to adapt to new domains, tasks, or instructions using relatively small amounts of labeled data while preserving the capabilities acquired during pre-training~\citep{wei2021finetuned}. In particular, instruction fine-tuning adapts a pre-trained model using a collection of datasets formatted as natural language instructions. Extensive work has demonstrated that this procedure consistently improves generalization to unseen tasks~\citep{wei2022finetunedlanguagemodelszeroshot, ouyang2022traininglanguagemodelsfollow, chung2022scalinginstructionfinetunedlanguagemodels, sanh2022multitaskpromptedtrainingenables, lambert2024tulu}. While~\cite{lambert2024tulu} perform a full-parameter fine-tune of the base model, we diverge from their approach and instead employ low-rank adaptation (LoRA)~\citep{hu2022lora}. In short, given a pre-trained weight matrix $W_0 \in \mathbb{R} ^{d_{\mathrm{out}} \times d_{\mathrm{in}}}$, LoRA freezes $W_0$ and parameterizes the update as a low-rank decomposition
\[
W = W_0 + BA,
\qquad
B \in \mathbb{R}^{d_{\mathrm{out}} \times r},
\quad
A \in \mathbb{R}^{r \times d_{\mathrm{in}}},
\quad
r \ll \min\{d_{\mathrm{out}},d_{\mathrm{in}}\},
\]
so that only the factors $A$ and $B$ are
learned.\footnote{We set $r = 16$ in all experiments.} We adopt LoRA over full-parameter fine-tuning for three reasons. First, \cite{biderman2024lora} demonstrate that LoRA acts as a strong implicit regularizer: it adapts the model to new tasks while incurring substantially less representational drift than full fine-tuning, thereby mitigating catastrophic forgetting of previously acquired capabilities. Second, LoRA is significantly more compute-efficient: with rank $r = 16$, the number of trainable parameters per layer reduces from $d_{\mathrm{in}} \times d_{\mathrm{out}}$ to $r(d_{\mathrm{in}} + d_{\mathrm{out}})$, a reduction of over two orders of magnitude for typical hidden dimensions. Finally, prior work shows that LoRA can match the performance of full-parameter fine-tuning within 1–2\% when properly tuned, while preserving its primary advantages of reduced memory usage and faster training \citep{lee2026learning, zhao2024lora}.

\begin{table}[ht]
\caption{Datasets used across training stages. The FineWeb corpus is indexed by publication timestamp; all fine-tuning datasets are temporally filtered to remove references to real-world events beyond the target cutoff (Appendix~\ref{app:leakage}).}
\label{tab:training_datasets}
\small
\begin{tabular}{llrr}
\toprule
\textbf{Stage} & \textbf{Dataset}~\cite{penedo2024fineweb} & \textbf{Tokens} & \textbf{Filtered} \\
\midrule
PT & \texttt{HuggingFaceFW/fineweb} & 170B / 1T & -- \\
\midrule
\textbf{Stage} & \textbf{Dataset}~\cite{lambert2024tulu, xu2024magpie}
 & \textbf{Examples} & \textbf{Filtered} \\
\midrule
\multirow{6}{*}{SFT}
 & \texttt{ai2-adapt-dev/evol\_codealpaca\_heval\_decontaminated} & 106,790 & 100,114 \\
 & \texttt{ai2-adapt-dev/personahub\_code\_v2\_34999} & 34,943 & 34,748 \\
 & \texttt{ai2-adapt-dev/tulu\_v3.9\_open\_math\_2\_gsm8k\_50k} & 50,000 & 49,772 \\
 & \texttt{ai2-adapt-dev/numinamath\_tir\_math\_decontaminated} & 64,191 & 63,910 \\
 & \texttt{ai2-adapt-dev/personahub\_ifdata\_manual\_seed\_v3\_29980} & 29,827 & 25,293 \\
 & \texttt{argilla/ifeval-like-data} & 456,304 & 386,584 \\
\bottomrule
\end{tabular}
\end{table}

\textbf{Data.} Table~\ref{tab:training_datasets} summarizes the datasets used in each training stage. While the FineWeb corpus is already indexed with publication timestamps allowing us to enforce a temporal cutoff directly. The version of FineWeb available at the time of writing spans 2013 through June 2025. Since we release year-end checkpoints, our latest released checkpoint is dated December 2024, the end of the last complete calendar year covered by the corpus. The remaining fine-tuning datasets required additional curation to ensure that no instruction-response pairs reference real-world events beyond the target time horizon. We describe our temporal filtering procedure in Appendix~\ref{app:leakage}. In addition, we remove examples exceeding the model's context length and filter out non-English instances to maintain a consistent linguistic distribution across all training stages. We cap the Argilla/IFEval-like~\citep{xu2024magpie} data at 270,000 examples to maintain an approximate balance whereby half of the dataset comprises coding and mathematical problems, while the remaining half focuses on rigorous adherence to user instructions.

\textbf{Evaluation.} We rely on the widely used~\cite{eval-harness} library to evaluate both our models and the benchmark models, supporting the reproducibility of our results.\footnote{All benchmarks for our PIT-1.5B\_2024 and PIT-4B\_2024 models are computed in fp32 precision.} In short,~\cite{eval-harness} provides the research community with a common library that offers standardized implementations for evaluating LLM performance across a wide range of end-to-end benchmarks.

\section{Results}
\label{sec:results}
\subsection{Common Sense Reasoning \& Language Comprehension}
\begin{figure}
    \centering
    \includegraphics[width=1.0\linewidth]{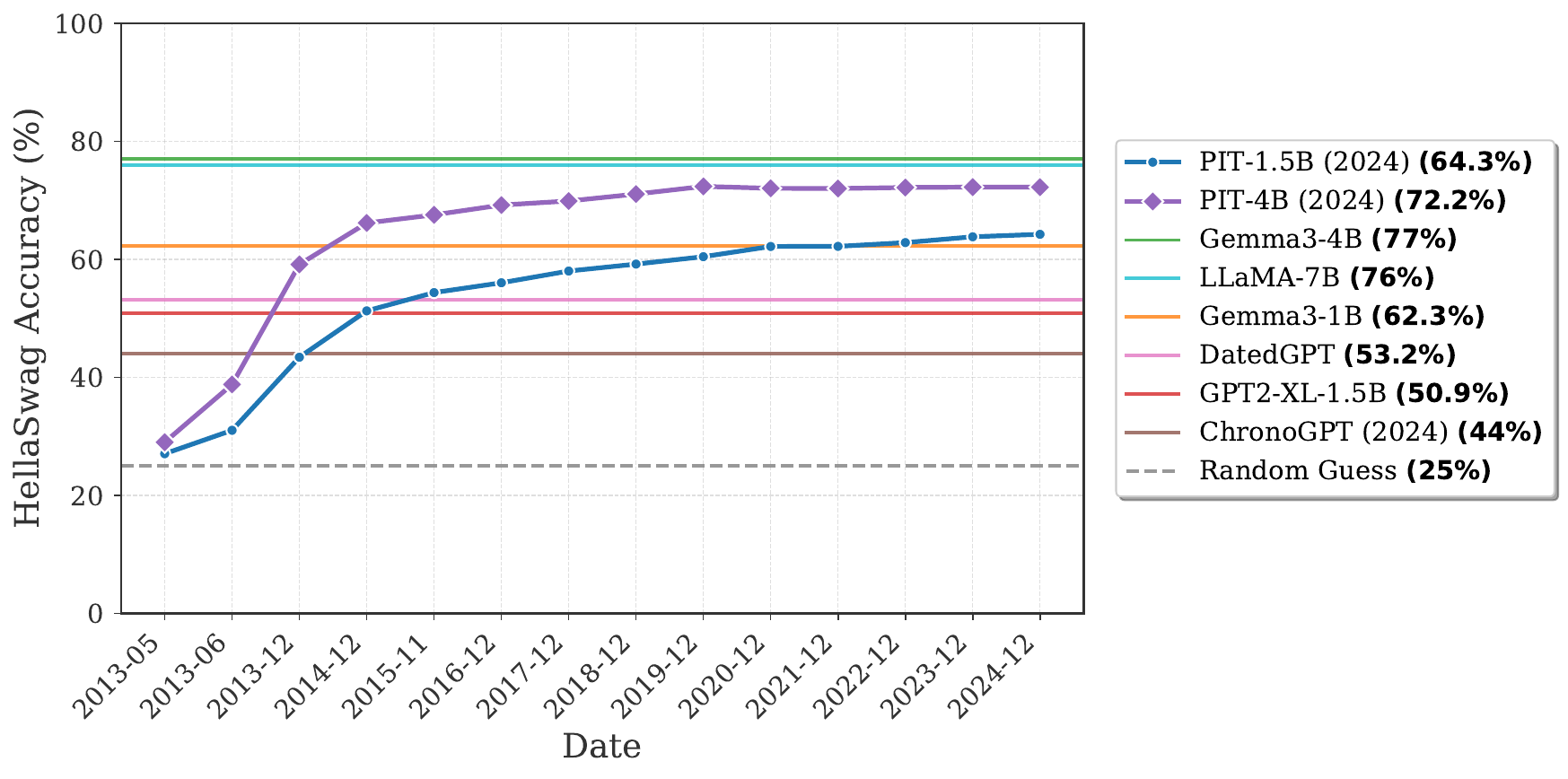}
    \caption{HellaSwag accuracy (\%) over time for our PIT-1.5B (170B tokens) and PIT-4B (1T tokens) models trained on chronologically ordered FineWeb data. Horizontal lines denote published baselines: Gemma3-4B (77\%)~\cite{team2025gemma}, Llama-7B (76\%)~\cite{touvron2023llama}, GPT2-XL (50.9\%)~\cite{radford2019language} (as reported in~\cite{wu2024lamini}), Gemma3-1B (62.3\%)~\cite{team2025gemma}, ChronoGPT 2024 (44\%)~\cite{he2025chronologically}, DatedGPT (53.2\%)~\cite{yan2026datedgpt}, and random guess (25\%).}
    \label{fig:placeholder}
\end{figure}

\begin{table}[htb]
\centering
\caption{Zero-shot accuracy (\%) on standard common sense reasoning benchmarks.}
\label{tab:pt_results}
\resizebox{\textwidth}{!}{
\begin{tabular}{lcccccccc}
\toprule
Model & BoolQ & PIQA & HellaSwag & WinoGrande & ARC-easy & ARC-chal. & OBQA & Avg. \\
\midrule
DatedGPT\_2024 & & 70.5 & 53.2 & & 52.0 & 34.7 & & 52.6\\
ChronoGPT\_2024
& 60.4 & 66.5 & 43.9 & 54.9 & 52.5 & 29.5 & 34.8 & 48.9 \\
PIT-1.5B\_2024 (Ours)
& 61.9 & 76.1 & 64.3 & 59.4 & 49.5 & 30.4 & 34.8 & 53.8 \\
PIT-4B\_2024 (Ours)
& \textbf{63.0} & \textbf{78.9} & \textbf{72.2} & \textbf{64.2} & \textbf{54.4} & \textbf{35.1} & \textbf{39.0} & \textbf{58.1} \\

\midrule
Gemma-3-1B
& 66.4 & 74.8 & 62.0 & 58.9 & 72.2 & 38.3 & 37.0 & 58.5 \\
Gemma-3-4B
& 79.0 & 80.0 & 76.0 & 69.5 & 81.8 & 54.9 & 43.0 & 69.2 \\
LLaMA-7B
& 76.8 & 79.7 & 76.0 & 69.6 & 72.1 & 44.3 & 44.4 & 66.1 \\
\bottomrule
\end{tabular}
}
\end{table}

Following~\cite{touvron2023llama}, we first evaluate our pre-trained models on common-sense reasoning tasks that assess whether language models have internalized broad world knowledge and can perform implicit reasoning beyond surface-level pattern matching. We report zero-shot accuracy on seven widely used benchmarks: BoolQ~\citep{clark2019boolq}, PIQA~\citep{bisk2020piqa}, 

HellaSwag~\citep{zellers2019hellaswag}, WinoGrande~\citep{sakaguchi2021winogrande}, ARC (Easy and Challenge)~\citep{clark2018arc}, and OpenBookQA~\citep{mihaylov2018openbookqa}.
All evaluations are conducted in the zero-shot setting, following standard practice in the language modeling literature.\footnote{The horizontal lines in Figure~\ref{fig:placeholder} show the values published by the respective authors, which do not all use the zero-shot setting: Gemma~3 reports 10-shot HellaSwag accuracy (77.2\% for Gemma-3-4B and 62.3\% for Gemma-3-1B)~\cite{team2025gemma}. Table~\ref{tab:pt_results} instead reports our own zero-shot evaluations with the LM Evaluation Harness~\cite{eval-harness}, which explains the small differences (e.g., 76.0\% vs.\ 77.2\% for Gemma-3-4B).}

Table~\ref{tab:pt_results} reports the results.\footnote{At the time of writing, we were unable to locate the DatedGPT weights and therefore report the numbers from the original paper~\cite{yan2026datedgpt}.} PIT-4B outperforms every other point-in-time LLM, including its 1.5B-parameter counterpart PIT-1.5B\_2024, ChronoGPT\_2024 \citep{he2025chronologically}, and DatedGPT~\cite{yan2026datedgpt}, with particularly large gains on HellaSwag (+28.3pp and +19pp over ChronoGPT\_2024 and DatedGPT\_2024, respectively). More importantly, PIT-4B closes much of the gap to LLaMA-7B and Gemma-3-4B---models trained without any temporal restriction and, in the case of LLaMA-7B, with nearly twice the parameter count. On PIQA and WinoGrande, the gap to LLaMA-7B and Gemma-3-4B is nearly negligible. These results demonstrate that chronological consistency need not come at the cost of strong language understanding.

\subsection{Instruction Following}
\label{sec:sft}
\begin{figure}
    \centering
    \includegraphics[width=0.75\linewidth]{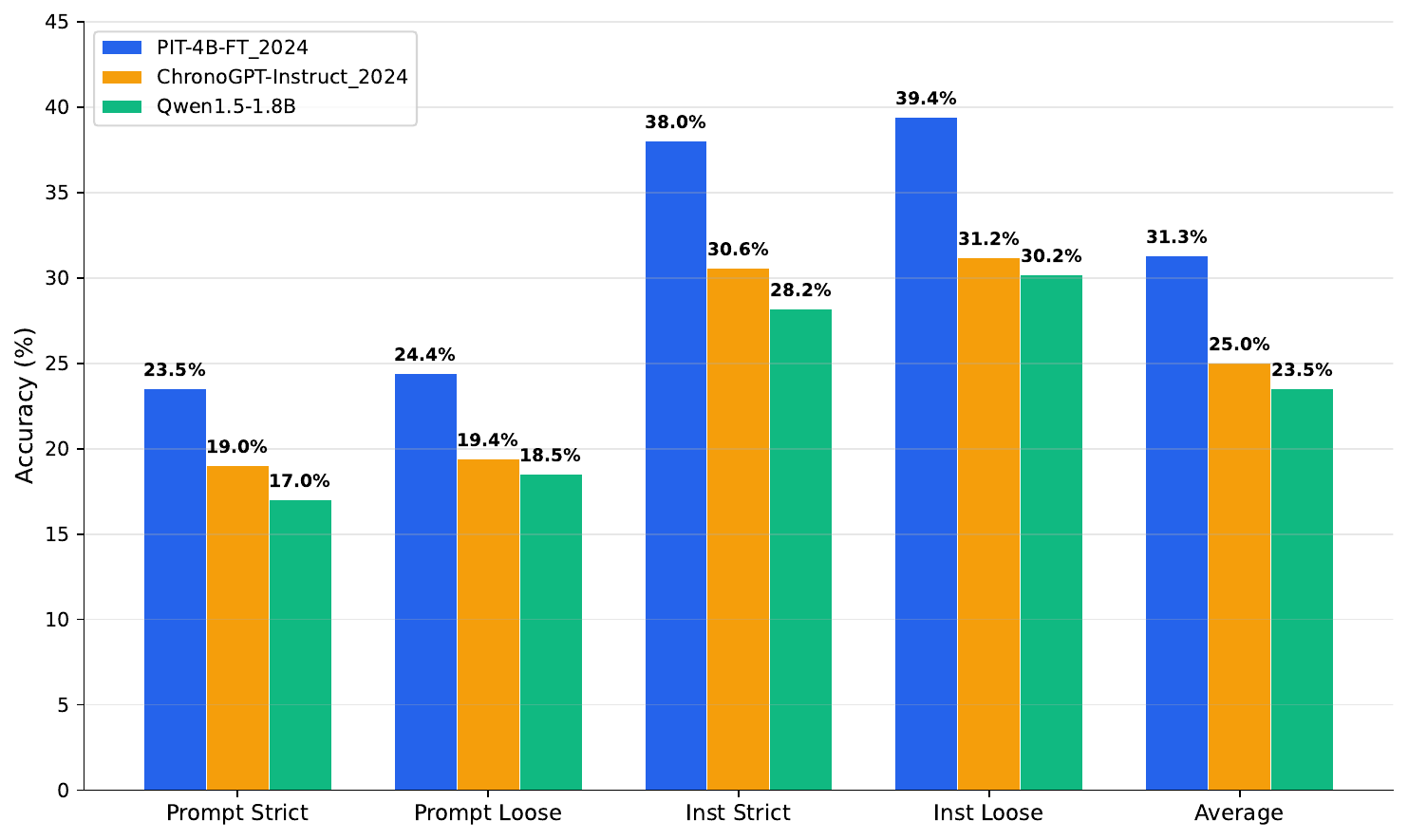}
    \caption{IFEval instruction-following accuracy (\%) for our fine-tuned PIT-4B-FT\_2024 compared with the point-in-time baseline ChronoGPT-Instruct\_2024~\cite{he2025instruction} and the temporally unrestricted model Qwen1.5-1.8B~\cite{bai2023qwen}.}
    \label{fig:IF}
\end{figure}

\textbf{Evaluation with IFEval.} A common approach to evaluating instruction-tuned models is to use an LLM as a judge~\citep{liu2023g, fu2023gptscore, chiang2023can}. For instance, \cite{he2025instruction} evaluate their instruction-tuned ChronoGPT using AlpacaEval~\citep{dubois2024length}, reporting length-controlled win rates against Qwen1.5-1.8B-Chat~\citep{bai2023qwentechnicalreport}. However, LLM-as-a-judge metrics are reference-free and suffer from well-documented biases: self-bias, where the judge prefers its own outputs~\citep{liu2023llms}; positional bias, where the judge favors responses based on presentation order~\citep{wang2023large}; and verbosity bias, where the judge prefers longer answers~\citep{zheng2023judging}. \cite{zheng2024cheating} demonstrate that a ``null model'' outputting constant, prompt-irrelevant responses can achieve top scores on AlpacaEval, Arena-Hard-Auto~\citep{li2024crowdsourced}, and MT-Bench~\citep{bai2024mt}. We refer the reader to~\cite{ye2024justice} for a comprehensive discussion of these failure modes.

To avoid these pitfalls, we evaluate instruction following using IFEval~\citep{zhou2023instruction}, a benchmark that tests compliance with verifiable constraints---length limits, required keywords, formatting rules---checked programmatically rather than by an LLM judge. Figure~\ref{fig:IF} reports IFEval instruction-following accuracy for PIT-4B-FT, ChronoGPT-Instruct, and Qwen1.5-1.8B on four metrics: prompt-strict, prompt-loose, instruction-strict, and instruction-loose. PIT-4B-FT attains the best performance on all four metrics, averaging 31.3\% versus 25.0\% and 23.5\% for ChronoGPT-Instruct and Qwen1.5-1.8B~\cite{bai2023qwen}, respectively, showing that our larger, fine-tuned architecture delivers materially stronger instruction-following than both the smaller chronologically consistent baseline and Qwen trained on temporally unrestricted data. Table~\ref{tab:ifeval-vintage} in Appendix~\ref{app:ifeval} provides additional details by comparing the average IFEval performance of vintage base and FT models (2013--2020) with that of later checkpoints (2021--2024).

Every model is evaluated using the default \texttt{IFEval} settings to maximize reproducibility, following the best practices of the public leaderboard\footnote{\url{https://huggingface.co/spaces/open-llm-leaderboard/open_llm_leaderboard\#/}}: temperature $=0$, maximum generation tokens $=1280$, and \texttt{do\_sample=False}.

\section{The Economic Evaluation of Embeddings}
\label{sec:economic-eval}
A key motivation for point-in-time language models is their use in financial applications where lookahead bias can invalidate results. Following~\cite{chen2022expected} and~\cite{didisheim2026inefficient}, we evaluate the economic value of our models by extracting news embeddings $E_t \in \mathbb{R}^{N_t\times d_h}$ and using them to predict stock returns $R_{t+1} \in \mathbb{R}^{N_t}$, where $N_t\in \mathbb{N_+}$ denotes the number of stocks available at time $t$ and $d_h$ is the embedding dimension. Because our models are trained exclusively on text available at each point in time, any return predictability we document is free of training leakage by construction, \emph{meaning the portfolios conditioned on stock news are fully tradable}.

\emph{To quantify economic performance}, we evaluate portfolios using the Sharpe ratio (SR)~\cite{sharpe1998sharpe}, the standard economic metric of risk-adjusted return, which measures average return per unit of volatility. This provides a natural analogue to downstream evaluation in machine learning: a model is useful insofar as it produces signals that translate into high SR portfolios out of sample.

\textbf{Data and Point-in-Time implementation.} We use the Dow Jones Newswire dataset which spans the period from June 1979 to May 2020. Because the earliest available point-in-time (PIT) language model is timestamped December 2013, we report out-of-sample performance for 2014 through May 2020 only. Starting in 2014, we implement a strictly out-of-sample embedding procedure: for each calendar year $t$, we use the PIT model available at the end of the previous year (December $t-1$) to embed all articles published during year $t$. For example, the December 2013 model is used for 2014 articles, the December 2014 model for 2015 articles, and so on. This rolling scheme ensures that embeddings are constructed using only information available at each point in time, thereby avoiding look-ahead bias.

\textbf{Full-sample benchmark models.}
In addition to the point-in-time (PIT) specifications, we consider two benchmark models based on the final checkpoints of our PIT models, denoted \textit{PIT-4B-full} and \textit{PIT-4B-FT-full}. Concretely, these correspond to the last available checkpoints obtained after training on the full sample. These models are fixed over time and therefore incorporate information from the entire sample, including future observations relative to each evaluation date. Consequently, they do not satisfy the strict temporal constraints imposed on the PIT models.

\textbf{Embedding Construction.} 
Let $d_h$ denote the dimension of the final hidden representation of the language model (e.g., $d_h = 4096$ for PIT-4B). For an article $i$ about stock $s$ published on day $d$, let $\mathbf{h}_{d,s}^{(i)} \in \mathbb{R}^{d_h}$ be the final-layer hidden state of the last token. We define the article embedding as
\[
\mathbf{e}_{d,s}^{(i)} := \mathbf{h}_{d,s}^{(i)}.
\]

We aggregate embeddings in two steps. First, we construct daily stock-level embeddings:
\[
\bar{\mathbf{e}}_{d,s}
=
\frac{1}{N_{d,s}}
\sum_{i=1}^{N_{d,s}}
\mathbf{e}_{d,s}^{(i)},
\]
where $N_{d,s}$ is the number of articles about stock $s$ on day $d.$ 
Second, we aggregate to the monthly level:
\[
\tilde{\mathbf{e}}_{t,s}
=
\frac{1}{|\mathcal{D}_{t,s}|}
\sum_{d \in \mathcal{D}_{t,s}}
\bar{\mathbf{e}}_{d,s},
\]
where $\mathcal{D}_{t,s}$ denotes the set of days in month $t$ where stock $s$ has at least one article.

\textbf{News-Only Information \& Anisotropy.} To isolate the component of news embeddings that is orthogonal to well-known cross-sectional predictors, we residualize $\tilde{\mathbf{e}}_{t,s}$ with respect to a rich set of firm characteristics. Specifically, we use the characteristics compiled by \cite{jensen2023there} (JKP), denoted $S_{t,s} \in \mathbb{R}^{K}$, which summarize standard risk factors and anomalies widely used in the asset pricing literature, and we apply the same data pre-processing as in~\cite{didisheim2024apt}.
Then, for each month $t$, we run the cross-sectional regression
\[
\tilde{\mathbf{e}}_{t,s} =\alpha + \beta S_{t,s} + \hat{\boldsymbol{\epsilon}}_{t,s}, \qquad \beta \in \mathbb{R}^{d_h\times K}, \qquad \alpha \in \mathbb{R}^{d_h}
\]
and use the residual $\hat{\boldsymbol{\epsilon}}_{t,s}$ as the characteristic-adjusted embedding. Thus, we work with monthly observations $E_t = [\hat{\epsilon}_{t,1}, \ldots, \hat{\epsilon}_{t,N_t}]^{\top} \in \mathbb{R}^{N_t \times d_h}$. It is worth noting that this process not only extracts ``pure'' news information but also solves the well-known anisotropy problem that generative model embeddings usually suffer from; see~\cite{ethayarajh2019contextual, gao2019representation}.

\textbf{Portfolio Construction.} We construct $d_h$ base portfolios from the embeddings $E_t\in \mathbb{R}^{N_t\times d_h}$ and denote their returns at time $t$ by
\[
\boldsymbol{F}_{t} = (F_{t,1}, \dots, F_{t,d_h})^\top \in \mathbb{R}^{d_h},
\quad \text{where} \quad
F_{t,i} = E_{t-1,i}^{\top} R_t \in \mathbb{R}.
\]
Here $E_{t-1,i}\in\mathbb{R}^{N_{t-1}}$ is the $i$-th column of $E_{t-1}$, i.e., the $i$-th embedding coordinate across stocks. Each column thus acts as a cross-sectional trading signal that assigns a weight to every stock, and $F_{t,i}$ is the realized return of the portfolio managed by that signal, so no first-stage return regression is needed. However, nothing guarantees that these base portfolios deliver high risk-adjusted returns. We therefore combine these portfolios using penalized Maximum Sharpe Ratio Regression (MSRR)~\cite{kelly2023financial}. This step can be interpreted as a regularized mean-variance optimization over the span of the base portfolios, in the spirit of Markowitz mean-variance portfolio choice \citep{markowitz1952modern} and the regression representation of the tangency portfolio in \cite{britten1999sampling}. We estimate MSRR weights using an expanding window $\mathcal{T}$ with an initial estimation period of 12 months.
\begin{equation}
\label{eq:MSRR_clean}
\hat{\boldsymbol{\lambda}}_t(z)
=
\arg\min_{\boldsymbol{\lambda} \in \mathbb{R}^{d_h}}
\frac{1}{\mathcal{T}}\sum_{u=1}^{\mathcal{T}}
\left(1 - \boldsymbol{\lambda}^\top \mathbf{F}_{u} \right)^2
+ z \|\boldsymbol{\lambda}\|_2^2.
\end{equation}
The resulting tradable portfolio is
\[
\boldsymbol{\pi}_t^{\text{PIT}}(z)
= E_t \hat{\boldsymbol{\lambda}}_t(z)
\in \mathbb{R}^{N_t},
\]
and the portfolio return is
\[
r_{t+1}^{\text{PIT}} = \boldsymbol{\pi}_t^{\text{PIT}}(z)^{\top} R_{t+1}.
\]

Hence, the final PIT portfolio is obtained by first constructing embedding-weighted base portfolios and then choosing the maximum-Sharpe combination of those portfolios.
At each rebalancing date, we solve the MSRR problem over a grid of shrinkage parameters $z \in \mathcal{Z}$. We rescale each $z$ according to
\[
z_{\text{eff}}
=
\frac{d_h}{\mathcal{T}}\, z \cdot \frac{1}{d_h}\,\mathrm{tr}\!\left(\frac{1}{\mathcal{T}}\dot{\mathbf{F}}_{\mathcal{T}}^\top \dot{\mathbf{F}}_{\mathcal{T}}\right), \qquad \mathcal{Z} = \{10^{-6}, \dots, 10^{-1}, 1, 2, 5, 10, 10^2\}
\]
where $\dot{\mathbf{F}}_{\mathcal{T}}
= (\boldsymbol{F}_1, \dots, \boldsymbol{F}_{\mathcal{T}})^\top
\in \mathbb{R}^{\mathcal{T}\times d_h} $, and
\(\mathrm{tr}(\dot{\mathbf F}_{\mathcal{T}}^\top \dot{\mathbf F}_{\mathcal{T}}/\mathcal{T}{})\) is the total second moment of the base portfolio returns estimated over the rolling window~\cite{chernov2025test}. 

For each value of $z$, we compute the returns of the strategy. We then rescale them to have a target volatility of $10\%$ and build an equally weighted portfolio of the strategy across the grid of $z.$ All performance statistics are computed over the post-December 2013 period, which constitutes the true out-of-sample evaluation.

\begin{figure}
    \centering
    \includegraphics[width=0.55\linewidth]{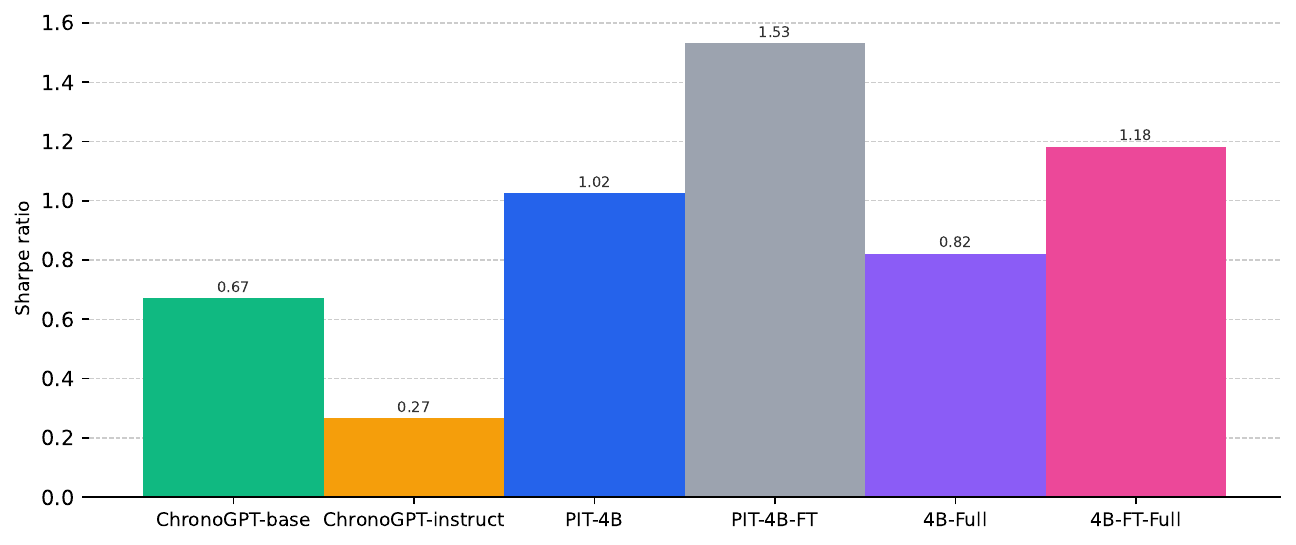}
    \caption{Out-of-sample annualized Sharpe ratio for each model.}
    \label{fig:sharpe-ratio}
\end{figure}
Figure \ref{fig:sharpe-ratio} reports out-of-sample annualized Sharpe ratios of the MSRR portfolios across model variants. Three main findings emerge:

\begin{enumerate}
    \item \textbf{Point-in-time models produce economically meaningful signals} even under strict temporal constraints. PIT models deliver consistently positive Sharpe ratios, substantially outperforming smaller ChronoGPT variants, which often exhibit weaker performance. Importantly, their performance remains close to that of the corresponding full-sample benchmarks, indicating that enforcing strict temporal validity does not materially reduce economic significance. 

    \item \textbf{Scaling substantially improves performance}. Moving from smaller ChronoGPT models to PIT-4B variants leads to large and systematic increases in Sharpe ratios. For example, Sharpe ratios increase from roughly 0.7 for ChronoGPT-base to around 1.0--1.5 for PIT-4B models. This indicates that model scale plays a central role in extracting economically valuable signals.

    \item \textbf{Fine-tuning further enhances performance at scale}. The fine-tuned PIT-4B model (PIT-4B-FT) outperforms its base counterpart. This suggests that fine-tuning might improve robustness and generalization of the extracted signals across the cross-section.

\end{enumerate}

\paragraph{Spanning test.}
To further understand how the signals generated from the models differ, we run a spanning test.
Following the standard empirical asset pricing literature, we evaluate outperformance via pairwise factor-spanning regressions
\[
r_{Target,t} = \alpha + \beta r_{benchmark,t} + \epsilon_t
\]
estimated on monthly returns, and report annualized alphas with annualized standard errors $SE(\alpha)$ in Table~\ref{tab:pairwise-benchmark}. The result is best read as a spanning statement, since alpha is asymmetric and both directions are informative. Our models earn large and highly significant alphas against the ChronoGPT baselines: PIT-4B-FT earns $13.07\%$ against ChronoGPT-base and $13.65\%$ against ChronoGPT-instruct, and it earns $11.41\%$ against our own non-fine-tuned PIT-4B. In the reverse direction, neither ChronoGPT variant earns a significant alpha against any of our models. The portfolios built from our embeddings span those built from ChronoGPT, but not the reverse, which is the standard criterion for concluding that one set of signals subsumes another.
\begin{table}[h!]
  \caption{Pairwise spanning regressions. Each cell reports the annualized alpha from regressing the monthly portfolio returns of the column model on those of the row model, with \cite{newey1987simple} standard errors in parentheses. Significance levels: $^{*}p<0.10$, $^{**}p<0.05$, and $^{***}p<0.01$.}
  \label{tab:pairwise-benchmark}
  \centering
  \footnotesize
  \setlength{\tabcolsep}{4pt}
  \begin{tabular}{l rrrrr r@{\hspace{12pt}}}
    \toprule
    \textbf{Benchmark Model (Row)} & (1) & (2) & (3) & (4) & (5) & (6) \\
    \midrule
    \textbf{(1) ChronoGPT-base}
      & ---
      & $-$1.42\%
      & 4.54\%\sym{**}
      & \textbf{13.07\%}\sym{***}
      & 4.39\%
      & 4.99\%\sym{**} \\
      &
      & (2.88\%) & (2.09\%) & (3.68\%) & (3.33\%) & (2.05\%) \\
    \addlinespace
    \textbf{(2) ChronoGPT-instruct}
      & 3.77\%
      & ---
      & 4.93\%\sym{**}
      & \textbf{13.65\%}\sym{***}
      & 6.88\%\sym{**}
      & 6.15\%\sym{***} \\
      & (2.34\%) &
      & (1.99\%) & (3.63\%) & (3.51\%) & (2.14\%) \\
    \addlinespace
    \textbf{(3) PIT-4B}
      & 3.14\%
      & $-$0.81\%
      & ---
      & \textbf{11.41\%}\sym{***}
      & 3.66\%
      & 4.73\%\sym{**} \\
      & (3.03\%) & (3.56\%) &
      & (3.66\%) & (3.70\%) & (2.19\%) \\
    \addlinespace
    \textbf{(4) PIT-4B-FT}
      & 3.36\%
      & 0.65\%
      & 3.20\%
      & ---
      & 5.07\%
      & 4.49\%\sym{*} \\
      & (3.24\%) & (3.95\%) & (2.22\%) &
      & (4.24\%) & (2.36\%) \\
    \addlinespace
    \textbf{(5) PIT-4B-full}
      & 1.58\%
      & $-$1.07\%
      & 3.51\%\sym{*}
      & \textbf{12.46\%}\sym{***}
      & ---
      & 4.34\%\sym{**} \\
      & (2.55\%) & (3.31\%) & (1.95\%) & (3.67\%) &
      & (1.99\%) \\
    \addlinespace
    \textbf{(6) PIT-4B-FT-full}
      & 1.28\%
      & $-$0.59\%
      & 3.40\%
      & \textbf{11.34\%}\sym{***}
      & 2.41\%
      & --- \\
      & (2.87\%) & (3.69\%) & (2.11\%) & (3.74\%) & (3.65\%)
      & \\
    \bottomrule
  \end{tabular}
\end{table}

\section{Conclusion}
\label{sec:conclusion}
This paper examines whether enforcing strict temporal validity in language model training requires sacrificing either general language-model performance or the downstream economic value of text embeddings. Our results suggest that it does not. By scaling point-in-time pre-training to 1 trillion chronologically filtered tokens and 4 billion parameters, we obtain models that substantially outperform prior chronologically consistent baselines and approach the performance of strong open-weight models trained on temporally unrestricted corpora. The main message is that much of the apparent cost of eliminating lookahead bias reflects a scale deficit rather than an inherent limitation of the point-in-time paradigm.

We show this both in standard NLP evaluations and in a financially meaningful downstream application.
On common-sense reasoning and language comprehension benchmarks, PIT-4B closes a large share of the gap to Gemma-3-4B and LLaMA-7B despite operating under a strict chronological constraint. After instruction fine-tuning with LoRA on temporally filtered data, the model also exhibits materially improved instruction-following performance on IFEval, indicating that temporal consistency can be preserved throughout the post-training pipeline. In an asset-pricing application, embeddings extracted from our point-in-time models generate economically meaningful out-of-sample return predictability, with larger models delivering stronger Sharpe ratios. These findings highlight the practical value of point-in-time models in empirical settings where leakage can otherwise invalidate inference.

More broadly, our results suggest that point-in-time language modeling is now a viable foundation for research in finance, economics, and the social sciences. In many of these settings, real-world policy interventions are costly to implement, slow to evaluate, politically constrained, and often ethically or administratively infeasible to randomize. Historically grounded in silico experiments can therefore play an important role in the design of complex social systems. But such exercises are credible only when the model’s information set respects chronology: once temporal leakage is present, ex post information contaminates the historical decision environment and undermines both causal and counterfactual interpretation. Point-in-time LLMs provide a natural computational remedy by enforcing chronological consistency in the training corpus. Our results show that researchers no longer need to choose between temporal validity and competitive language-model performance. By releasing the full training, filtering, and evaluation pipeline, we aim to make chronologically consistent language modeling reproducible and easier to adopt in applications where causal interpretation, historical fidelity, and backtest integrity are essential.

\textbf{Limitations.} Several limitations remain. First, although scaling substantially narrows the performance gap between PIT models and temporally unrestricted models, it does not eliminate it completely, particularly on some reasoning-intensive tasks. Second, our asset-pricing application focuses on a single benchmark dataset and a single portfolio construction framework; broader evidence across tasks and domains remains an important direction for future work. Finally, our finding that PIT models deliver greater economic gains than full-sample models is consistent with recent work in economics and finance, including \cite{didisheim2026inefficient} and \cite{he2025chronologically}. A plausible explanation is that language models are trained for next-token prediction rather than return prediction: exposure to future text may alter the embedding geometry without necessarily producing representations that are more useful for forecasting returns. Nevertheless, full-sample embeddings remain temporally invalid because the embedding function was learned using information unavailable at the historical evaluation date. Developing a rigorous theoretical framework to explain these differences in economic performance is an important direction for future research.


\section{Acknowledgments}
The authors thank anonymous reviewers for their constructive feedback. Semyon Malamud gratefully acknowledges the financial support of the Swiss Finance Institute and the Swiss National Science Foundation, Grant 100018-228042. AQR Capital Management is a global investment management firm that may or may not apply similar investment techniques or methods of analysis as described herein. The views expressed here are those of the authors and not necessarily those of AQR. This work was supported by a grant from the Swiss National Supercomputing Centre (CSCS) under project ID lp46 on Alps. This work was supported under project ID abs25 as part of the Swiss AI Initiative, through a small grant from the ETH Domain and computational resources provided by the Swiss National Supercomputing Centre (CSCS) under the Alps infrastructure. This paper was written with assistance from Claude, an AI assistant by Anthropic. We thank Asaf Manela, Marko Stojanovic, Rocco Ventruto, and Yutong Yan for helpful comments.
\clearpage
\newpage
\bibliographystyle{plainnat}
\bibliography{bib}
\clearpage
\appendix

\section{Training Configuration}
\label{app:training-config}

 We train two GPT-2-style decoder-only transformer models on chronologically ordered FineWeb data. PIT-1.5B (released on Hugging Face as PIT-1B) has 52 layers, 12 attention heads, and an embedding dimension of 1,536 (head dim = 128), totaling approximately 1.5B parameters trained on 170B tokens. PIT-4B has 20 layers, 32 attention heads, and an embedding dimension of 4,096 (head dim = 128), totaling approximately 4.2B parameters trained on 1T tokens. Both models use a vocabulary of 50,304 tokens with a sequence length of 2,048. Training employs a hybrid optimizer: Muon (momentum = 0.95) for the transformer blocks and AdamW ($\beta_1$ = 0.9, $\beta_2$ = 0.95) for the language model head, with a peak learning rate of 0.0009 and no weight decay. The learning rate follows a trapezoidal schedule with linear warmdown over the final 50\% of training steps. We ran all experiments on NVIDIA GH200s and H100s, depending on queue and resource availability. PIT-4B achieved roughly 10,000 tokens per second per GPU, so the full pretraining run took roughly 28,000 GPU-hours.

\section{Removing Temporal Leakage}
\label{app:leakage}
Similar to~\cite{he2025instruction}, we employ gpt-5-nano to identify and remove temporally sensitive examples from the datasets used in the supervised fine-tuning stage. We present each example from the SFT datasets to gpt-5-nano, prepending the prompt illustrated in Figure~\ref{fig:temporal_prompt}, and retain only those examples classified as ``timeless"---namely, examples that do not reference real-world events, real individuals, specific dates, pop culture, or trending topics. This filtering step is designed to ensure that the training data remains invariant to temporal context, thereby preventing the model from internalizing ephemeral or potentially outdated knowledge during fine-tuning. Notably, we observe that the majority of examples in every dataset are naturally retained through this process, as instruction-following data predominantly comprises verifiable format constraints rather than world knowledge~\citep{zhou2023instruction}, while mathematical and coding problems are inherently grounded in abstract reasoning, formal logic, and counterfactual scenarios that are not subject to temporal drift.  

\begin{figure}[ht]
\centering
\fbox{\parbox{0.92\textwidth}{\small\ttfamily
You are a binary classifier. Determine if text contains
TIME-SENSITIVE FACTS that could become outdated or incorrect
over time.\\ \\
Output ONLY "0" (timeless) or "1" (time-aware).\\ \\
Output "1" ONLY if the text states facts that could BECOME
OUTDATED, such as:\\
-- Who currently holds a political office
   ("Biden is the president")\\
-- Recent or dated events
   ("The 2024 Olympics were held in Paris")\\
-- Current statistics, prices, or rankings
   ("Tesla stock is at \$X")\\
-- A real person doing something specific at a specific time
   ("Elon Musk announced X in 2024")\\
-- Laws, policies, or regulations tied to a specific time
   frame\\ \\
Output "0" if the text is:\\
-- Math problems or solutions (even with character names like
   "Alice" or "John")\\
-- Programming/coding tasks or tutorials (even mentioning real
   tools: Python, Java, NetBeans, Eclipse, AWS, Docker, etc.)\\
-- General educational content about real software, frameworks,
   or technologies\\
-- Generic advice or best practices (even about real
   products)\\
-- Hypothetical scenarios (even with human names)\\
-- Instruction-following tasks\\
-- Scientific facts that don't change
   ("water boils at 100C")\\
-- Abstract reasoning or logic puzzles\\ \\
KEY DISTINCTION: Mentioning a real tool, company, or product
by name does NOT make text time-aware. Only FACTUAL CLAIMS
THAT COULD BECOME OUTDATED do.\\ \\
Examples:\\
-- "Use NetBeans profiler for CPU analysis"
   \textrightarrow\ 0 (generic advice about a tool)\\
-- "Write a REST API using Flask and AWS Lambda"
   \textrightarrow\ 0 (coding tutorial)\\
-- "Google announced Gemini 2.0 in December 2024"
   \textrightarrow\ 1 (dated event)\\
-- "The president of the United States is Joe Biden"
   \textrightarrow\ 1 (will become outdated)\\
-- "Tesla's market cap exceeded \$1 trillion in 2024"
   \textrightarrow\ 1 (time-sensitive fact)\\
-- "Solve: 2x + 3 = 7"
   \textrightarrow\ 0 (math)\\
-- "Anna bought 5 apples at \$2 each"
   \textrightarrow\ 0 (hypothetical)\\
-- "Python's GIL prevents true multithreading"
   \textrightarrow\ 0 (technical fact, stable)\\
-- "React 18 introduced concurrent rendering"
   \textrightarrow\ 0 (historical tech fact, won't change)\\
-- "As of 2024, React is the most popular framework"
   \textrightarrow\ 1 (ranking changes over time)
}}
\caption{Prompt used for temporal leakage classification.}
\label{fig:temporal_prompt}
\end{figure}
\clearpage
\section{Mean IFEval Performance: Checkpoints comparison}\label{app:ifeval}
We compare the IFEval performance of the base model, PIT-4B, and its instruction-tuned counterpart, PIT-4B-FT, across two groups of checkpoints: early vintages (2013--2020) and later vintages (2021--2024). Table~\ref{tab:ifeval-vintage} reports scores averaged across checkpoints within each period, separately for the base and fine-tuned models.
\begin{table}[h!]
  \caption{Mean IFEval performance: early (2013--2020) vs.\ later (2021--2024) checkpoints.}
  \label{tab:ifeval-vintage}
  \centering
  \small
  \begin{tabular}{l rrr}
    \toprule
    \textbf{Metric} & 2013--2020 & 2021--2024 & ($\Delta$, pp) \\
    \midrule
    Base Prompt S & 17.14\% & 15.40\% & $-$1.74 \\
    Base Prompt L & 18.31\% & 16.43\% & $-$1.89 \\
    Base Inst S   & 29.70\% & 28.02\% & $-$1.68 \\
    Base Inst L   & 31.25\% & 29.27\% & $-$1.98 \\
    \midrule
    FT Prompt S   & 20.05\% & 23.05\% & $+$3.00 \\
    FT Prompt L   & 21.16\% & 23.62\% & $+$2.46 \\
    FT Inst S     & 33.42\% & 37.85\% & $+$4.43 \\
    FT Inst L     & 34.60\% & 38.80\% & $+$4.20 \\
    \bottomrule
  \end{tabular}
\end{table}
\newpage
\clearpage
\section*{NeurIPS Paper Checklist}
\begin{enumerate}

\item {\bf Claims}
    \item[] Question: Do the main claims made in the abstract and introduction accurately reflect the paper's contributions and scope?
    \item[] Answer: \answerYes{}
    \item[] Justification: In the abstract, we highlight the important problem of look-ahead bias in empirical applications of large language models in economics, policy design, and the social sciences. We make three claims. First, we advance the state of the art in point-in-time LLMs, as shown in Section~\ref{sec:results}. Second, we provide a look-ahead-bias-free evaluation based on portfolios constructed from embeddings~\ref{sec:economic-eval}. Finally, the code is fully available online \href{https://github.com/DjoFE2021/Point-In-Time-LLM}{GitHub}.

\item {\bf Limitations}
    \item[] Question: Does the paper discuss the limitations of the work performed by the authors?
    \item[] Answer: \answerYes{}
    \item[] Justification: See Section~\ref{sec:conclusion}.

\item {\bf Theory assumptions and proofs}
    \item[] Question: For each theoretical result, does the paper provide the full set of assumptions and a complete (and correct) proof?
    \item[] Answer: \answerNA{} 
    \item[] Justification: We pre-train large language models using the well-known next-token prediction objective~\cite{radford2018improving, bengio2003neural} and then fine-tune them in a supervised manner. Finally, we conduct a standard economic evaluation of the embeddings generated by these models~\cite{sharpe1998sharpe}.

    \item {\bf Experimental result reproducibility}
    \item[] Question: Does the paper fully disclose all the information needed to reproduce the main experimental results of the paper to the extent that it affects the main claims and/or conclusions of the paper (regardless of whether the code and data are provided or not)?
    \item[] Answer: \answerYes{} 
    \item[] Justification: Yes, the code is fully available online, and the algorithms are fully documented in the main text; see the \href{https://github.com/DjoFE2021/Point-In-Time-LLM}{GitHub} repository.

\item {\bf Open access to data and code}
    \item[] Question: Does the paper provide open access to the data and code, with sufficient instructions to faithfully reproduce the main experimental results, as described in supplemental material?
    \item[] Answer: \answerYes{} 
    \item[] Justification: Yes, the code is fully available online via our \href{https://github.com/DjoFE2021/Point-In-Time-LLM}{GitHub} repository. The FineWeb~\cite{penedo2024fineweb} and Tulu-3~\cite{lambert2024tulu} datasets are publicly available on Hugging Face. Access to the JKP~\cite{jensen2023there} and Dow Jones datasets, however, may require a subscription, although such access is often available to academics at no cost.

\item {\bf Experimental setting/details}
    \item[] Question: Does the paper specify all the training and test details (e.g., data splits, hyperparameters, how they were chosen, type of optimizer) necessary to understand the results?
    \item[] Answer: \answerYes{} 
    \item[] Justification: See Section~\ref{sec:methodology} and Appendix~\ref{app:training-config}.

\item {\bf Experiment statistical significance}
    \item[] Question: Does the paper report error bars suitably and correctly defined or other appropriate information about the statistical significance of the experiments?
    \item[] Answer: \answerYes{} 
    \item[] Justification: For the economic evaluation, Table~\ref{tab:pairwise-benchmark} reports annualized alphas from pairwise spanning regressions estimated on monthly out-of-sample portfolio returns (January 2014 to May 2020), with Newey--West standard errors~\cite{newey1987simple}, which are robust to heteroskedasticity and autocorrelation, and significance levels ($^{*}p<0.10$, $^{**}p<0.05$, $^{***}p<0.01$). Since the embeddings and portfolio rules are deterministic once the model weights are fixed, these standard errors capture the sampling variability of monthly returns over the evaluation period. The language-model benchmarks (Table~\ref{tab:pt_results} and Figures~\ref{fig:placeholder} and~\ref{fig:IF}) are single deterministic evaluations with the LM Evaluation Harness~\cite{eval-harness} (zero-shot; greedy decoding for IFEval) and, as is standard, are reported without error bars.

\item {\bf Experiments compute resources}
    \item[] Question: For each experiment, does the paper provide sufficient information on the computer resources (type of compute workers, memory, time of execution) needed to reproduce the experiments?
    \item[] Answer: \answerYes{} 
    \item[] Justification: See Appendix~\ref{app:training-config}.
    
\item {\bf Code of ethics}
    \item[] Question: Does the research conducted in the paper conform, in every respect, with the NeurIPS Code of Ethics \url{https://neurips.cc/public/EthicsGuidelines}?
    \item[] Answer: \answerYes{} 
    \item[] Justification: No violations.

\item {\bf Broader impacts}
    \item[] Question: Does the paper discuss both potential positive societal impacts and negative societal impacts of the work performed?
    \item[] Answer: \answerYes{} 
    \item[] Justification: Yes, in Section~\ref{sec:intro}, Section~\ref{sec:literature}, and Section~\ref{sec:conclusion}, we explain the importance of point-in-time LLMs in fields other than computer science, such as economics and policymaking. In these fields, it is essential to construct counterfactuals and conduct in silico experiments without temporal leakage.
    
\item {\bf Safeguards}
    \item[] Question: Does the paper describe safeguards that have been put in place for responsible release of data or models that have a high risk for misuse (e.g., pre-trained language models, image generators, or scraped datasets)?
    \item[] Answer: \answerNA{}.
    \item[] Justification: No risk of misuse.

\item {\bf Licenses for existing assets}
    \item[] Question: Are the creators or original owners of assets (e.g., code, data, models), used in the paper, properly credited and are the license and terms of use explicitly mentioned and properly respected?
    \item[] Answer: \answerYes{} 
    \item[] Justification: To the best of our knowledge, we have cited and credited all creators and owners of the existing assets used in this work.

\item {\bf New assets}
    \item[] Question: Are new assets introduced in the paper well documented and is the documentation provided alongside the assets?
    \item[] Answer:  \answerYes{} 
    \item[] Justification: Yes, the code is available at our \href{https://github.com/DjoFE2021/Point-In-Time-LLM}{GitHub} repository. The model weights and training data are publicly available on \href{https://huggingface.co/Diamegs}{Hugging Face}, with a model card for each checkpoint.

\item {\bf Crowdsourcing and research with human subjects}
    \item[] Question: For crowdsourcing experiments and research with human subjects, does the paper include the full text of instructions given to participants and screenshots, if applicable, as well as details about compensation (if any)? 
    \item[] Answer: \answerNA{}.
    \item[] Justification: NA.

\item {\bf Institutional review board (IRB) approvals or equivalent for research with human subjects}
    \item[] Question: Does the paper describe potential risks incurred by study participants, whether such risks were disclosed to the subjects, and whether Institutional Review Board (IRB) approvals (or an equivalent approval/review based on the requirements of your country or institution) were obtained?
    \item[] Answer: \answerNA{}.
    \item[] Justification: NA.

\item {\bf Declaration of LLM usage}
    \item[] Question: Does the paper describe the usage of LLMs if it is an important, original, or non-standard component of the core methods in this research? Note that if the LLM is used only for writing, editing, or formatting purposes and does \emph{not} impact the core methodology, scientific rigor, or originality of the research, declaration is not required.
    \item[] Answer: \answerYes{}.
    \item[] Justification: We used the OpenAI model gpt-5-nano to identify and filter temporally dependent instruction-following observations from~\cite{lambert2024tulu} and~\cite{xu2024magpie}, as described in Section~\ref{sec:methodology} and Appendix~\ref{app:leakage}.

\end{enumerate}
\end{document}